\documentclass[10pt,twocolumn,letterpaper]{article}

\usepackage{cvpr}              
\definecolor{cvprblue}{rgb}{0.21,0.49,0.74}
\usepackage[pagebackref,breaklinks,colorlinks,allcolors=cvprblue]{hyperref}

\def\paperID{MTF07} 
\def\confName{CVPR}
\def\confYear{2026}

\title{Evaluating the Feasibility of Inferring Dietary Behavior Change Receptivity from Egocentric Images of Eating Environment}

\author{Long Li\\
The University of Alabama\\
Tuscaloosa, AL, USA\\
\and
Yuning Huang\\
Purdue University\\
West Lafayette, IN, USA\\
\and
Heather A. Eicher-Miller\\
Purdue University\\
West Lafayette, IN, USA\\
\and
J.Graham Thomas\\
Brown University\\
Providence, RI, USA\\
\and
Fengqing Zhu\\
Purdue University\\
West Lafayette, IN, USA\\
\and
Edward Sazonov\\
The University of Alabama\\
Tuscaloosa, AL, USA\\
}

\begin{document}
\maketitle
\begin{abstract}
Accurately assessing dietary behavior change receptivity is essential for designing effective just-in-time adaptive interventions (JITAIs) that promote healthier eating habits. However, self-report–based assessment of behavior change receptivity is sparse and delayed, limiting its practical use in continuous monitoring. To explore whether passive sensing may help address this challenge, this study conducts a pilot investigation of inferring participants’ self-reported behavior change receptivity from egocentric eating images collected by a wearable camera. We use pilot data obtained from free-living eating episodes using the Automatic Ingestion Monitor v2 (AIM-2). The data included egocentric image sequences captured during eating and paired with responses to questions assessing specific dimensions of behavior change receptivity (awareness, interaction capability, and motivation). To examine whether visual information contained any relevancy to these responses, we evaluated a transfer-learning-assisted framework that combines a pre-trained Contrastive Language–Image Pre-Training (CLIP) vision encoder with a lightweight transformer classifier. The model processes eating episode image sequences to extract potential semantic and temporal cues related to behavior change receptivity. Preliminary experimental results show promising improvements over simple baseline models for behavior change receptivity indicators. These early findings suggest that egocentric eating episode images may contain cues related to dietary behavior change receptivity, and warrant further investigation with larger and more comprehensive datasets.
\end{abstract}    
\section{Introduction}
Eating behavior is an important determinant of long-term health, and many chronic conditions can be prevented by improving dietary habits \cite{fard2023compliance, walsh2011importance}. Digital health technologies aim to support these improvements by providing timely guidance during daily life. A key requirement for such systems is the ability to understand the user’s real-time eating context and their willingness to adopt behavioral changes \cite{abernethy2022promise}. In most current systems, this information is collected through self-reports. Although effective in controlled studies, self-reports are difficult to obtain continuously in real-world settings because they rely on user compliance and retrospective recall \cite{dakin2025exploring, sazonov2010energetics}. As a result, it remains challenging to monitor eating behavior and dietary behavior change receptivity at scale.

Wearable egocentric cameras offer a potential solution \cite{li2025extralightweight}. They capture visual information from the user’s perspective during everyday activities and can do so without interrupting behavior \cite{huang2025progressive}. Prior research has used such data for tasks like detecting eating episodes, recognizing foods, and estimating portion size \cite{bell2020automatic, hossain2020realtime, jobarteh2020development}. These methods focus mainly on observable aspects of behavior. Much less work examines whether visual data can provide insight into higher-level constructs such as attentional state or willingness to modify eating habits \cite{ghosh2024integrated}. Progress in this direction is further limited by the scarcity of real-world data that pair wearable eating episode images with meaningful behavioral labels \cite{guo2025zeroshot}. As a result, it remains uncertain whether visual cues from eating episode images might offer any useful signal for estimating subjective behavioral information.

In this study, we examine whether dietary behavior change receptivity can be inferred from eating episode image sequences captured in free-living conditions. We define dietary behavior change receptivity as an individual's momentary attentional availability, capacity to engage with the intervention, and motivation to modify eating behaviors in response to dietary guidance. We analyze pilot data collected using a egocentric wearable camera system called AIM-2 \cite{doulah2021automatic} that records egocentric images during eating event. At the same time, participants provide self-reported dietary behavior change receptivity responses upon review of the images after the eating events. This combination of visual observations and self-reports enables an initial examination of whether any subjective behavior-related signals are reflected in real-world eating episode images.

We formulate this as a supervised learning task at the event level. To address the challenge of limited labeled data and the visual diversity of free-living images, we develop a transfer learning framework based on a pre-trained CLIP vision encoder. CLIP generates semantic visual representations that generalize well across diverse scenes, while a lightweight transformer-based classifier predicts the corresponding dietary behavior change receptivity indicators. This approach enables the model to leverage large-scale pre-training for a preliminary exploration of whether visual information may relate to self-reported dietary behavior change receptivity in a small-data setting.

The main contributions of this paper are summarized as follows:
\begin{itemize}
    \item First, we conduct an initial investigation of whether dietary behavior change receptivity can be approximately estimated from real-world eating episode image sequences collected by an egocentric wearable camera.
    \item Second, we propose a transfer learning--based approach that leverages a pre-trained CLIP vision encoder and a lightweight classifier to explore the extent to which dietary behavior change receptivity can be inferred from egocentric eating episode images.
\end{itemize}

The remainder of this paper is organized as follows. Section II reviews related work that motivates our study. Section III describes the pilot data collection protocol. Section IV presents the proposed transfer learning--based classifier in detail. Section V reports the experimental setup and results follow by discussion. Section VI concludes the paper and outlines directions for future research.
\section{Related Work}
This section reviews prior work that motivates our study. We summarize existing datasets and methods for vision-based dietary and behavior analysis, highlighting the gap in data that link free-living egocentric eating observations with behavior-related (subjective) labels. We then briefly introduce two technical building blocks used in our approach: transfer learning for robust semantic feature extraction and Transformer architectures for modeling temporal patterns in eating episodes.

\subsection{Related Datasets for Dietary and Behavior Analysis}
A significant body of research in vision-based dietary assessment has been built upon datasets of static, third-person food images, such as Food-101 \cite{bossard2014food}, UECFood-256 \cite{kawano2015automatic}, and other curated food recognition datasets (e.g., \cite{mao2021visual}). While foundational for food recognition, these datasets lack the crucial real-world context of an eating episode, such as the environment, eating progression, and the user’s interaction with food. As a result, they offer limited utility for understanding eating behavior in free-living conditions.

To incorporate such context, a second generation of datasets has emerged from egocentric (first-person) vision. Large-scale datasets like EPIC-KITCHENS \cite{damen2018scaling} and Ego4D \cite{grauman2022ego4d} provide extensive video of daily activities, including eating. However, annotations in these benchmarks primarily focus on physical action recognition (i.e., what the user is doing) rather than behavioral or psychological factors that influence eating. While these datasets are valuable for activity modeling, they do not include self-reported indicators of attention, intention, or motivation.

Our research also builds on prior work developing the Automatic Ingestion Monitor (AIM) platform \cite{farooq2016novel}. AIM and related wearable sensing systems have been pivotal for passively detecting eating episodes and ingestive microstructure, including chewing and swallowing events \cite{sazonov2010automatic, lopezmeyer2010detection, doulah2021automatic}. These studies provide insight into when and how eating occurs, but they focus on observable behaviors and do not include user-reported behavioral or motivational information needed to study constructs such as dietary behavior change receptivity.

Beyond dietary assessment, recent efforts in egocentric vision have begun exploring inference of internal states from first-person recordings. For example, egoEMOTION links egocentric video with dense self-reports of emotion and personality, illustrating that visual data can reflect aspects of internal experience \cite{jammot2025egoemotion}. This methodological direction suggests the potential feasibility of relating egocentric visual cues to non-physical behavioral attributes.

Collectively, existing efforts do not address behavior-related constructs specific to dietary behavior change receptivity. This gap motivates our pilot study, which examines whether dietary behavior change receptivity may be reflected in free-living egocentric eating episode images paired with self-reported responses.

\subsection{User Eating Behavior Assessment}
Current approaches for assessing eating behavior and motivation rely heavily on self-report instruments, including food diaries, 24-hour recalls, and Ecological Momentary Assessment (EMA) \cite{shiffman2008ecological, wrzus2023ecological}. While EMA can reduce recall bias by capturing data closer to the moment of experience, self-report methods still impose substantial participant burden, often resulting in sparse, delayed, and potentially inaccurate measurements. This limitation reduces their practicality for continuous monitoring and timely decision-making in just-in-time adaptive interventions (JITAIs) \cite{moses2023investigating}.

To reduce burden, research has increasingly shifted toward passive assessment using wearable sensors, largely focusing on quantifying physical eating behaviors. Examples include wrist-worn inertial sensors to detect hand-to-mouth gestures \cite{dong2014detecting} and acoustic sensing to detect chewing and swallowing \cite{sazonov2010automatic, lopezmeyer2010detection}. Vision-based systems (including AIM-style platforms) further enable passive capture of visual context related to when eating occurs and what foods are consumed \cite{doulah2021automatic, hossain2020realtime, hossain2024enhancing, huang2024automatic}.

However, an effective JITAI must account not only for observable eating patterns, but also for the user’s readiness or willingness to engage with and act on an intervention. Motivated by this gap, we explore whether internal behavioral factors related to eating (e.g., receptivity to dietary behavior change) may be reflected, at least partially, in the same egocentric visual data commonly used for dietary assessment.

\subsection{Transfer Learning}
Training deep neural networks for specialized tasks such as inferring receptivity from eating episode images is challenged by limited labeled data and the high variability of free-living environments. A common strategy is transfer learning, where models pre-trained on large general-purpose datasets (e.g., ResNet \cite{he2016deep} pre-trained on ImageNet \cite{deng2009imagenet}) are adapted to downstream tasks.

In this work, we leverage CLIP \cite{radford2021learning}, which learns visual representations via large-scale image--text pretraining. Compared with models trained only on fixed image category labels, CLIP provides broadly transferable semantic features that can generalize across diverse scenes and objects encountered in free-living egocentric imagery.

We therefore use the pre-trained CLIP vision encoder as a feature extractor, converting each image into a semantic embedding. This allows the downstream classifier to focus on learning the mapping from these high-level visual concepts to behavior-related labels, without needing to relearn low-level visual primitives from scratch.

\subsection{Transformer Models for Temporal Behavior Modeling}
An eating episode is inherently temporal: cues relevant to receptivity may arise from food-related content, eating dynamics (e.g., pace and interruptions), and environmental context (e.g., eating alone vs.\ socially, at a desk vs.\ in front of a TV). Because these cues can be distributed across the full episode, a model must aggregate information over long temporal ranges to produce a single event-level prediction.

Traditional sequence models (e.g., RNNs) can struggle to capture long-range dependencies \cite{bengio1994learning}. We therefore employ a Transformer encoder \cite{vaswani2017attention}, whose self-attention mechanism enables global context aggregation across all frames in the episode. This design is well-suited for exploratory modeling of event-level receptivity, as it can integrate the full sequence of CLIP embeddings and selectively emphasize the most salient temporal and contextual cues when forming an episode-level representation.
\begin{figure}[t]
    \centering
    \includegraphics[width=0.75\linewidth]{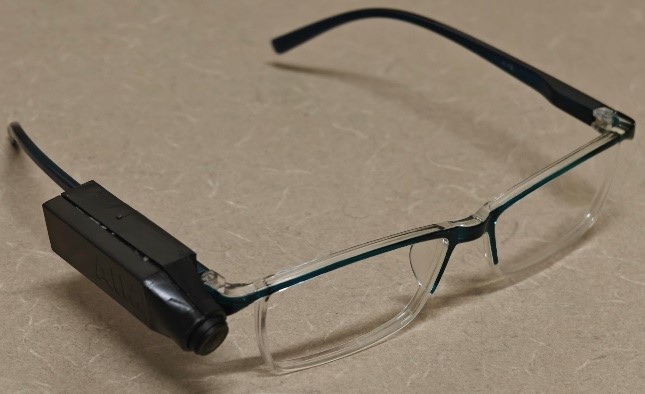}
    \caption{Automatic Ingestion Monitor v2 (AIM-2) wearable sensor.}
    \label{fig:aim2}
\end{figure}

\begin{figure*}[t]
    \centering
    \includegraphics[width=\textwidth]{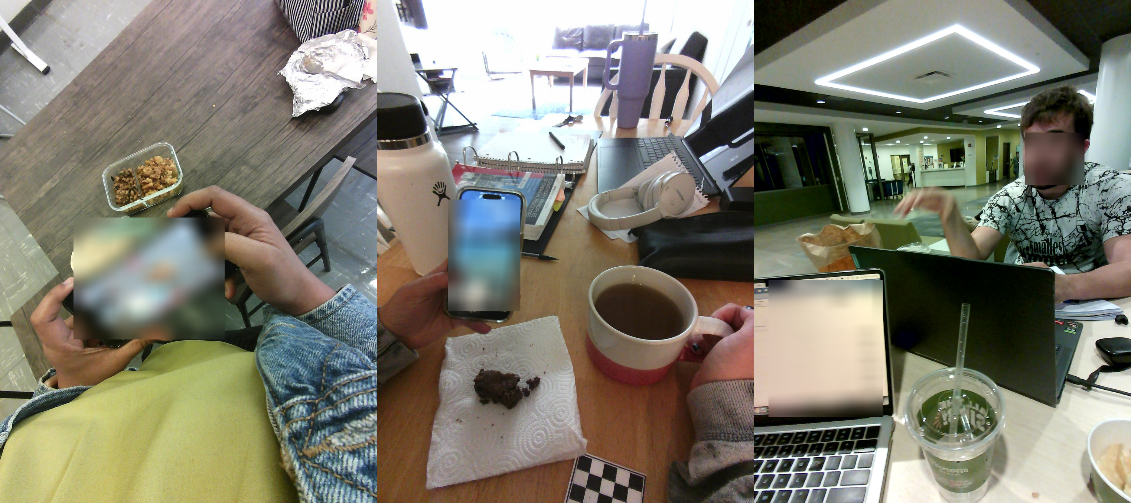}
    \caption{Examples of processed egocentric images in the pilot dataset.}
    \label{fig:processed_examples}
\end{figure*}
\section{Pilot Data Collection}
To evaluate the feasibility of predicting dietary behavior change receptivity from egocentric eating images, we collected pilot data in real-world, free-living eating environments. The study was conducted at Purdue University under IRB Protocol \#IRB-2022-270. Volunteer participants were instructed to wear the Automatic Ingestion Monitor v2 (AIM-2) for one full day \cite{doulah2021automatic}.

AIM-2 is a wearable dietary monitoring system that integrates a small, outward-facing egocentric camera with onboard ingestive sensing \cite{doulah2021automatic}. The device automatically detects eating events using built-in sensors and triggers image capture during those periods. When eating is detected, AIM-2 records egocentric images at approximately 10-second intervals, providing a visual stream of the user’s eating episodes without requiring any manual input. The system is designed for unobtrusive, day-long wear in naturalistic environments, enabling passive and ecological data collection.

After data collection, participants reviewed the captured images on a computer screen and were allowed to delete images that raised privacy concerns (e.g., images containing identifiable faces, bystanders, or digital screens). Participants then used custom software to answer questions about their receptivity to dietary behavior change for each eating episode. The software asked six questions, each rated on a 5-point Likert scale (1 = very unlikely, 2 = unlikely, 3 = neutral, 4 = likely, 5 = very likely). These questions assessed the participant’s ability to notice or interact with phone prompts during the eating episodes, as well as their perceived likelihood or motivation to change aspects of their eating behavior based on such prompts:
\begin{itemize}
    \item \textbf{Q1:} While eating, I would be able to \textbf{NOTICE A MESSAGE} sent to my phone encouraging me to change my eating behavior.
    \item \textbf{Q2:} While eating, I would be able to \textbf{OPEN AND INTERACT} with a message sent to my phone about changing my eating behavior.
    \item \textbf{Q3:} How likely is it that you could make changes to the \textbf{TYPES OF FOODS} that you ate after getting a phone message when you were eating?
    \item \textbf{Q4:} How likely is it that you could make changes to \textbf{HOW MUCH YOU ATE} after getting a phone message when you were eating?
    \item \textbf{Q5:} How likely is it that you could make changes to \textbf{HOW YOU ATE} (e.g., eating more slowly) after getting a phone message when you were eating?
    \item \textbf{Q6:} \textbf{HOW MOTIVATED} would you be to make changes to your eating behavior after getting a phone message when you were eating?
\end{itemize}

Data were collected from 78 participants. For each participant, the pilot dataset contains: (1) a participant ID, (2) the start and end times of each eating episode, (3) responses to the six predefined questions, and (4) all images captured during that eating episode. Across all participants, the dataset includes 17{,}239 images, with an average of 221 images per participant. A total of 173 eating episodes were recorded, corresponding to approximately 2.2 eating episodes per participant. All raw images were processed to remove privacy-sensitive content, such as human faces and digital screens, following privacy-preserving practices for egocentric wearable cameras \cite{li2025extralightweight, huang2025progressive}. Representative processed images are shown in Fig.~\ref{fig:processed_examples}, and the AIM-2 device is illustrated in Fig.~\ref{fig:aim2}.
\section{Method}
\label{sec:method}
\begin{figure*}[t]
    \centering
    \includegraphics[width=0.8\textwidth]{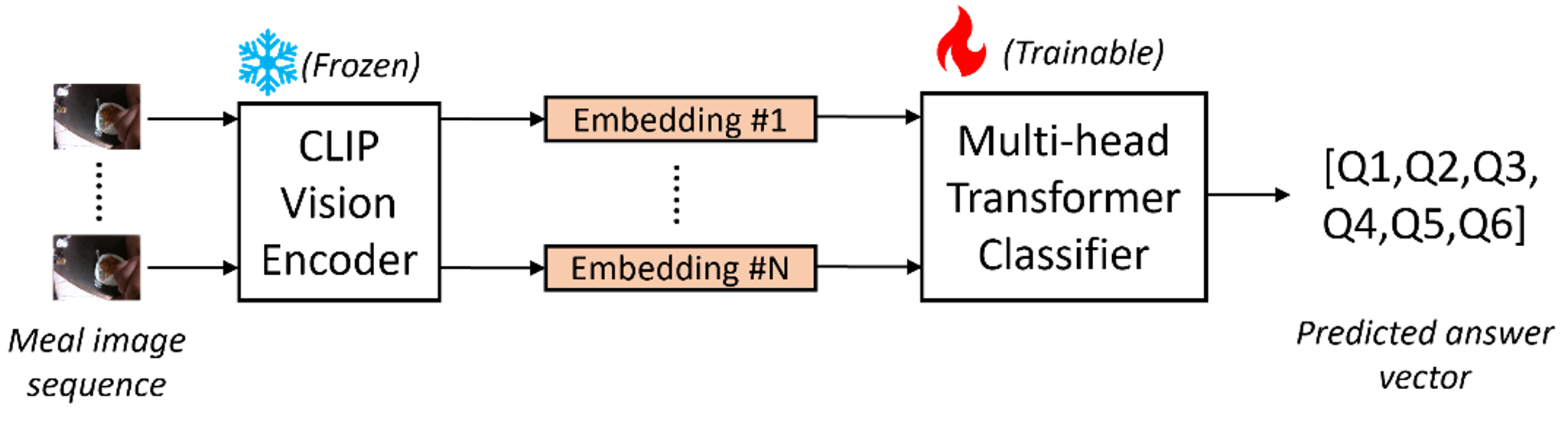}
    \caption{Overview of the proposed transfer learning framework.}
    \label{fig:method}
\end{figure*}
To predict dietary behavior change receptivity for each eating episode using only the egocentric images captured during that eating episode, we need to address the challenges of limited dataset size and high visual variability in free-living environments. For this, we adopt a transfer learning framework that combines a pre-trained CLIP vision encoder \cite{radford2021learning} with a lightweight transformer-based classifier \cite{vaswani2017attention}. The overall processing pipeline is illustrated in Fig.~\ref{fig:method}. Each eating episode images are first converted into a sequence of high-level visual embeddings. This sequence is then modeled as a temporal signal to produce a single event-level representation, which is finally mapped to the dietary behavior change receptivity outputs.

\subsection{Feature extraction using CLIP}
As illustrated in Fig.~\ref{fig:method}, we form the eating episode images into a sequence. For each eating episode, all images whose timestamps fall between the user self-reported start and end times are selected and ordered chronologically. Because the duration of eating episodes varies across participants, the number of available frames can differ substantially. To ensure consistent processing and training in batch, we define a maximum sequence length of $T=100$ frames. If an eating episode has more than $T$ frames, only the first $T$ images are used. If an eating episode has fewer than $T$ frames, zero-padding is applied to reach length $T$.

Once we get the images sequence, we use the CLIP ViT-B/32 vision encoder as a fixed feature extractor to obtain semantically rich embeddings for each eating episode image. Before being passed to CLIP, each image undergoes the model’s standard preprocessing steps, including resizing and normalization. The encoder operates in a frozen feature-extraction mode and produces a high-level embedding for each frame. These embeddings capture semantic properties of the eating scene, such as objects, utensils, food items, and aspects of the environment.

\subsection{Lightweight transformer-based classifier}
After the feature extraction process, the next step is to model the temporal relationships that unfold throughout the eating episode. Although individual frames provide useful information, many behavioral cues emerge across the sequence as the eating episode progresses. To capture these dynamics, we employ a lightweight transformer encoder.

The transformer encoder processes the entire sequence using self-attention, allowing it to integrate information from all frames and identify temporal patterns relevant to dietary behavior change receptivity outputs. The extracted sequence-level embeddings are passed through six independent prediction heads, each corresponding to one of the six questions collected from the participant. Each head is a small feed-forward network that outputs a classification result for the corresponding question. By sharing the sequence representation across these prediction heads, the model simultaneously produces exploratory estimates of all six dietary behavior change receptivity indicators associated with an eating episode.

\subsection{Experimental setting}
The key parameters used in all experiments are summarized in Table~\ref{tab:params}. The CLIP encoder produces 512-dimensional embeddings for each frame, and each eating episode image sequence is limited to a maximum of 100 frames. The transformer classifier is configured with two encoder layers, four attention heads, and a hidden dimension of 128. Training is performed using a batch size of 16 and a learning rate of $1\times10^{-4}$. These settings were kept consistent across all evaluation conditions.

\begin{table}[t]
\centering
\caption{Key parameters in experiment.}
\label{tab:params}
\begin{tabular}{l l}
\hline
\textbf{Category} & \textbf{Setting / Value} \\
\hline
Embedding Dimension & 512 \\
Max Sequence Length & 100 \\
Batch size & 16 \\
Epochs & 100 \\
Learning Rate & $1\times10^{-4}$ \\
Train/Validation Split & 80\% / 20\% \\
Transformer Layers & 2 \\
Attention Heads & 4 \\
Hidden Dimension & 128 \\
\hline
\end{tabular}
\end{table}

To obtain a reliable estimate of model performance, we conducted 5-fold cross-validation with participant-level splitting. In this setup, data was partitioned such that all eating episodes belonging to a single participant were grouped exclusively into one fold. This guarantees that during each iteration of the cross-validation, a participant’s data appears solely in either the training set or the validation set, but never both.

In addition to predicting the original 5-class Likert responses, we also evaluated a simplified binary classification version of the task. To construct the binary labels, Likert responses of 1--3 were grouped into the ``not agree'' category, while responses of 4--5 were grouped into the ``agree'' category. This additional setting allows us to examine whether reducing label granularity may offer a more tractable prediction task for pilot data.

\subsection{Comparison baseline methods}
In our experiment, we compared the proposed method against two simple baseline classifiers. The first baseline is a random classifier, which assigns a label by sampling uniformly from the possible classes (for both the 5-class Likert and binary settings). This provides a lower-bound reference that reflects performance expected by chance.

The second baseline is a majority classifier, which always predicts the most frequent label observed in the training set. This baseline reflects the performance achievable by exploiting class imbalance without learning any specific patterns. Together, these baselines help assess whether the proposed method provides any incremental predictive value beyond chance-level or majority-frequency predictions.
\section{Results and Discussion}
\label{sec:results}

In this section, we report performance for each behavior-related question and compare the proposed method against baseline classifiers. The goal is to examine whether visual information from egocentric eating episode images provides any signal for estimating receptivity-related self-reports.

\subsection{Performance on Original Likert Responses (5-Class)}
\label{subsec:likert_results}

Table~\ref{tab:likert_acc} summarizes classification accuracy on the original 5-point Likert responses. Overall, the proposed CLIP+Transformer framework consistently outperforms the random baseline across all six questions, and exceeds the majority baseline on four of the six questions (Q2, Q3, Q5, Q6). While the majority classifier performs slightly better for Q1 and Q4, the proposed method achieves the best average accuracy across all indicators.

\begin{table}[t]
\centering
\caption{Classification accuracy comparison on original Likert responses (1--5).}
\label{tab:likert_acc}
\resizebox{\columnwidth}{!}{
\begin{tabular}{lccc}
\hline
\textbf{Question} & \textbf{Random (\%)} & \textbf{Majority (\%)} & \textbf{Proposed (\%)} \\
\hline
Q1 & 19.1 & 27.2 & 23.3 \\
Q2 & 21.4 & 35.8 & 46.7 \\
Q3 & 20.8 & 30.6 & 40.0 \\
Q4 & 22.5 & 39.3 & 30.0 \\
Q5 & 17.3 & 41.0 & 56.7 \\
Q6 & 22.0 & 30.1 & 43.3 \\
\hline
Average & 20.5 & 34.0 & 40.1 \\
\hline
\end{tabular}}
\end{table}
\subsection{Performance on Merged Binary Responses}
\label{subsec:binary_results}

In addition to predicting the original Likert responses, we evaluate a simplified binary classification task by merging Likert responses 1--3 as ``not agree'' and 4--5 as ``agree.'' As shown in Table~\ref{tab:binary_acc}, reducing label granularity yields substantially improved performance. The proposed method outperforms both baselines on five of the six questions, with accuracy reaching up to 85.7\% (Q1) and an average accuracy of 75.0\%.

\begin{table}[t]
\centering
\caption{Classification accuracy comparison on merged binary responses (0/1).}
\label{tab:binary_acc}
\resizebox{\columnwidth}{!}{
\begin{tabular}{lccc}
\hline
\textbf{Question} & \textbf{Random (\%)} & \textbf{Majority (\%)} & \textbf{Proposed (\%)} \\
\hline
Q1 & 50.9 & 68.2 & 85.7 \\
Q2 & 52.0 & 59.0 & 84.0 \\
Q3 & 54.3 & 71.1 & 63.6 \\
Q4 & 52.0 & 56.6 & 74.4 \\
Q5 & 53.8 & 60.1 & 80.0 \\
Q6 & 48.0 & 43.4 & 62.5 \\
\hline
Average & 51.8 & 59.7 & 75.0 \\
\hline
\end{tabular}}
\end{table}

\subsection{Discussion and Limitations}
\label{subsec:discussion}

These results provide initial evidence that dietary behavior change receptivity may be inferred from passive egocentric visual sensing in free-living conditions. The observed gains over chance and frequency-based baselines suggest that CLIP-derived semantic embeddings preserve task-relevant cues when transferred to this domain.

Several limitations should be acknowledged. First, label distributions are imbalanced for some questions, and potential technical confounders (e.g., image clarity and capture angle variations) were not explicitly modeled. Second, psychological constructs are influenced by internal factors that may not be fully visible in images. Given the pilot-scale dataset, these findings should be interpreted as an exploratory step; future work should expand the dataset, incorporate additional wearable sensing modalities, and explore improved temporal modeling strategies to better capture individual differences and within-episode dynamics.

Practical deployment also depends on user acceptance of egocentric cameras and privacy-preserving data handling. In this pilot study, participants could delete sensitive images, and the dataset was processed to remove faces and screens; however, real-world adoption may be limited by concerns about bystanders, social acceptability, and whether computation occurs on-device or offloaded to servers. These factors motivate future work on privacy-preserving and on-device processing, and on measuring user willingness to use such systems in everyday settings.
\section{Conclusion}

This study presented a feasibility investigation into passively predicting dietary behavior change receptivity via egocentric vision. We introduced a transfer learning framework that synergizes the semantic breadth of CLIP with the temporal modeling of Transformers to decode high-level behavioral constructs from eating episode sequences.

Our experimental results suggest that this approach is promising for exploring whether egocentric eating images contain cues associated with self-reported receptivity. While the pilot study shows consistent improvements over simple baselines in several questions, the findings should be interpreted as preliminary due to the limited sample size and label imbalance. Future work with larger datasets and complementary sensing modalities is needed to validate generalizability and better characterize the visual cues that drive predictions.
{
    \small
    \bibliographystyle{ieeenat_fullname}
    \bibliography{main}
}


\end{document}